%% file: main.tex
\documentclass[letterpaper, 10 pt, conference]{ieeeconf}  % Comment this line out if you need a4paper

\IEEEoverridecommandlockouts                              % This command is only needed if 
\usepackage{graphics} % for pdf, bitmapped graphics files
\usepackage{epsfig} % for postscript graphics files
\usepackage{mathptmx} % assumes new font selection scheme installed
\usepackage{times} % assumes new font selection scheme installed
\usepackage{amsmath} % assumes amsmath package installed
\usepackage{amssymb}  % assumes amsmath package installed
\usepackage{array}
\usepackage{booktabs}
\usepackage{gensymb}
\usepackage{multirow}
\usepackage{xcolor}
\usepackage{color,soul}
\usepackage[noadjust]{cite} % multiple citations are combined

\DeclareMathOperator*{\argmin}{arg\,min}

\title{\LARGE \bf
MLNav: Learning to Safely Navigate on Martian Terrains
}

\author{Shreyansh Daftry$^{1}$, Neil Abcouwer$^{1}$, Tyler Del Sesto$^{1}$, Siddarth Venkatraman$^{2}$\\
Jialin Song$^{3}$, Lucas Igel$^{4}$, Amos Byon$^{1}$, Ugo Rosolia$^{3}$, Yisong Yue$^{3}$ and Masahiro Ono$^{1}$% <-this % stops a space
\thanks{$^{1}$S. Daftry, N. Abcouwer, T. Del Sesto, A. Byon and M. Ono are with Jet Propulsion Laboratory, California Institute of Technology, USA. $^{2}$S. Venkatraman is with Manipal Institute of Technology, India. $^{3}$J. Song, U. Rosolia and Y. Yue are with California Institute of Technology, USA. $^{4}$L. Igel is with Massachusetts Institute of Technology, USA. Email: daftry@jpl.nasa.gov}
}

\begin{document}

\maketitle
\pagestyle{empty}

%%%%%%%%%%%%%%%%%%%%%%%%%%%%%%%%%%%%%%%%%%%%%%%%%%%%%%%%%%%%%%%%%%%%%%%%%%%%%%%%
\begin{abstract}
%In this paper we present \textit{MLNav} -- a general learning-enabled path planning framework for safety-critical and resource-limited systems working in complex environments, such as Mars rovers. 
%MLNav makes judicious use of machine learning to enhance the efficiency of path planning while fully respecting safety constraints.
%A single run of machine-learned search heuristics makes predictions on the feasibility for all path options in a search tree so that the path planner can quickly find feasible paths. A computationally expensive, model-based collision checker is run only on top-ranked paths to guarantee safety. 
%Experiments in a high-fidelity simulator using the real Martian terrain data collected by the Perseverance rover, as well as a large-scale Monte-Carlo simulation with synthetic terrains, demonstrate that MLNav provides statistically significant increase in drive success rate while needing a substantially reduced number of runs of collision checks. 
%Notably, the improvement on real Martian terrains was achieved by a model that is only trained with synthetic data. 

We present \textit{MLNav}, a learning-enhanced path planning framework for safety-critical and resource-limited systems operating in complex environments, such as rovers navigating on Mars. 
MLNav makes judicious use of machine learning to enhance the efficiency of path planning while fully respecting safety constraints.
In particular, the dominant computational cost in such safety-critical settings is running a model-based safety checker on the proposed paths.  Our learned search heuristic can simultaneously predict the feasibility for all path options in a single run, and the model-based safety checker is only invoked on the top-scoring paths.
We validate in high-fidelity simulations
using both real Martian terrain data collected by the Perseverance rover, as well as a suite of challenging synthetic terrains. Our experiments show that: (i) compared to the baseline ENav path planner on board the Perserverance rover, MLNav can provide a significant improvement in multiple key metrics, such as a 10x reduction in collision checks when navigating real Martian terrains, despite being trained with synthetic terrains; and (ii) MLNav can successfully navigate highly challenging terrains where the baseline ENav fails to find a feasible path before timing out.
\end{abstract}

%%%%%%%%%%%%%%%%%%%%%%%%%%%%%%%%%%%%%%%%%%%%%%%%%%%%%%%%%%%%%%%%%%%%%%%%%%%%%%%%
\section{Introduction}
\label{subsec:intro}

\subsection{Motivation}
\input{sec_introduction}

\subsection{Statement of Contribution}
\label{sec:contrib}
\input{sec_contribution}

%%%%%%%%%%%%%%%%%%%%%%%%%%%%%%%%%%%%%%%%%%%%%%%%%%%%%%%%%%%%%%%%%%%%%%%%%%%%%%%%
\section{Related Work}
\input{sec_related}
\label{subsec:related}

%%%%%%%%%%%%%%%%%%%%%%%%%%%%%%%%%%%%%%%%%%%%%%%%%%%%%%%%%%%%%%%%%%%%%%%%%%%%%%%%
\section{MLNav Framework}
\label{sec:approach}
\input{sec_approach}

%%%%%%%%%%%%%%%%%%%%%%%%%%%%%%%%%%%%%%%%%%%%%%%%%%%%%%%%%%%%%%%%%%%%%%%%%%%%%%%%
\section{Deployment for Mars Rover Navigation}
\label{sec:enav}
\input{sec_enav}

%%%%%%%%%%%%%%%%%%%%%%%%%%%%%%%%%%%%%%%%%%%%%%%%%%%%%%%%%%%%%%%%%%%%%%%%%%%%%%%%
\section{Performance Evaluation}
\label{sec:exp}
\input{sec_experiments}

%%%%%%%%%%%%%%%%%%%%%%%%%%%%%%%%%%%%%%%%%%%%%%%%%%%%%%%%%%%%%%%%%%%%%%%%%%%%%%%%
\section{MLNav on Real Mars Data}
\label{sec:mars_exp}
\input{sec_mars_experiment}

\section{Conclusion}
\input{sec_conclusion}

% \addtolength{\textheight}{-12cm}   % This command serves to balance the column lengths
                                  % on the last page of the document manually. It shortens
                                  % the textheight of the last page by a suitable amount.
                                  % This command does not take effect until the next page
                                  % so it should come on the page before the last. Make
                                  % sure that you do not shorten the textheight too much.

%%%%%%%%%%%%%%%%%%%%%%%%%%%%%%%%%%%%%%%%%%%%%%%%%%%%%%%%%%%%%%%%%%%%%%%%%%%%%%%%

%%%%%%%%%%%%%%%%%%%%%%%%%%%%%%%%%%%%%%%%%%%%%%%%%%%%%%%%%%%%%%%%%%%%%%%%%%%%%%%%

\section{Acknowledgement}
The research was carried out at the Jet Propulsion Laboratory, California Institute of Technology, and California Institute of Technology under a contract with the National Aeronautics and Space Administration, and supported in part by Raytheon. The authors would like to thank Olivier Toupet, Mitch Ingham and Ravi Lanka for valuable discussions and problem formulation, and the JPL Research and Technology Development (R\&TD) program for supporting this research.

%%%%%%%%%%%%%%%%%%%%%%%%%%%%%%%%%%%%%%%%%%%%%%%%%%%%%%%%%%%%%%%%%%%%%%%%%%%%%%%%

\bibliographystyle{IEEEtran}
\bibliography{egbib}

\end{document}

%% file: sec_introduction.tex
NASA's new rover \textit{Perseverance} successfully landed on Mars in February 2021 with a new autonomous driving algorithm called Enhanced AutoNav (ENav) \cite{mchenry2020mars}. ENav, a classic tree-based path planner at its core, brought substantial upgrades to its predecessor. As of the writing of this article, ENav has driven the rover for approximately half of the 2.2 km driving distance, a remarkable achievement given that Curiosity rover has been driven autonomously for only $\sim$6.4\% of its driving distance to date \cite{rankin2020curiosity}. However, ENav's ability is still far behind human rover drivers, particularly on complex terrains where it often fails to find a feasible path even if one exists. In many cases, ground operators must manually drive the rover through challenging scenarios before transferring control to ENav as a way to extend the drive distance beyond the visible terrain. The highly stringent safety requirements (zero-tolerance for mechanical damage or unrecoverable situations) and the limited on-board computational resource are the major roadblocks for further improvements of extraterrestrial autonomous driving. 

%As mobile robots break out of controlled laboratory settings and are increasingly deployed in a real and complex environment like Mars, planning modules have to balance multiple contextual aspects, including computing resources, responsiveness, system-level performance, and most importantly, safety. This presents a challenge to classic motion planning pipelines for two reasons. First, most of the existing work has tended to focus on the development of tractable planning algorithms with provable worst-case performance guarantees such as computational complexity \cite{canny1988complexity}, probabilistic completeness \cite{lavalle1998rapidly} or asymptotic optimality \cite{karaman2011sampling}. In contrast, analysis of the finite-time performance of these algorithms has received considerably less attention \cite{choudhury2018adaptive}. Second, traditional methods are heavily model-based (in the sense that it requires models of the robot, the world and hand-designed heuristic functions) \cite{smith2018learning} and hence take little advantage of the contextual structure of the environment; thereby having to typically plan every path from scratch. This has a substantial influence on the finite-time planning performance \cite{hsu1997path}.

As mobile robots break out of controlled laboratory settings and are increasingly deployed in a real and complex environment like Mars, planning modules have to balance multiple contextual aspects, including computing resources, responsiveness, system-level performance, and most importantly, safety. This presents a challenge to classic motion planning pipelines as their computational complexity increases exponentially with the dimensionality of the motion planning problem. Furthermore, traditional methods are heavily model-based (in the sense that it requires models of the robot, the world and hand-designed heuristic functions) and hence take little advantage of the contextual structure of the environment; thereby having to typically plan every path from scratch. This has a substantial influence on the finite-time planning performance \cite{hsu1997path}.

\begin{figure}[t]
    \centering
    \includegraphics[width=0.85\linewidth]{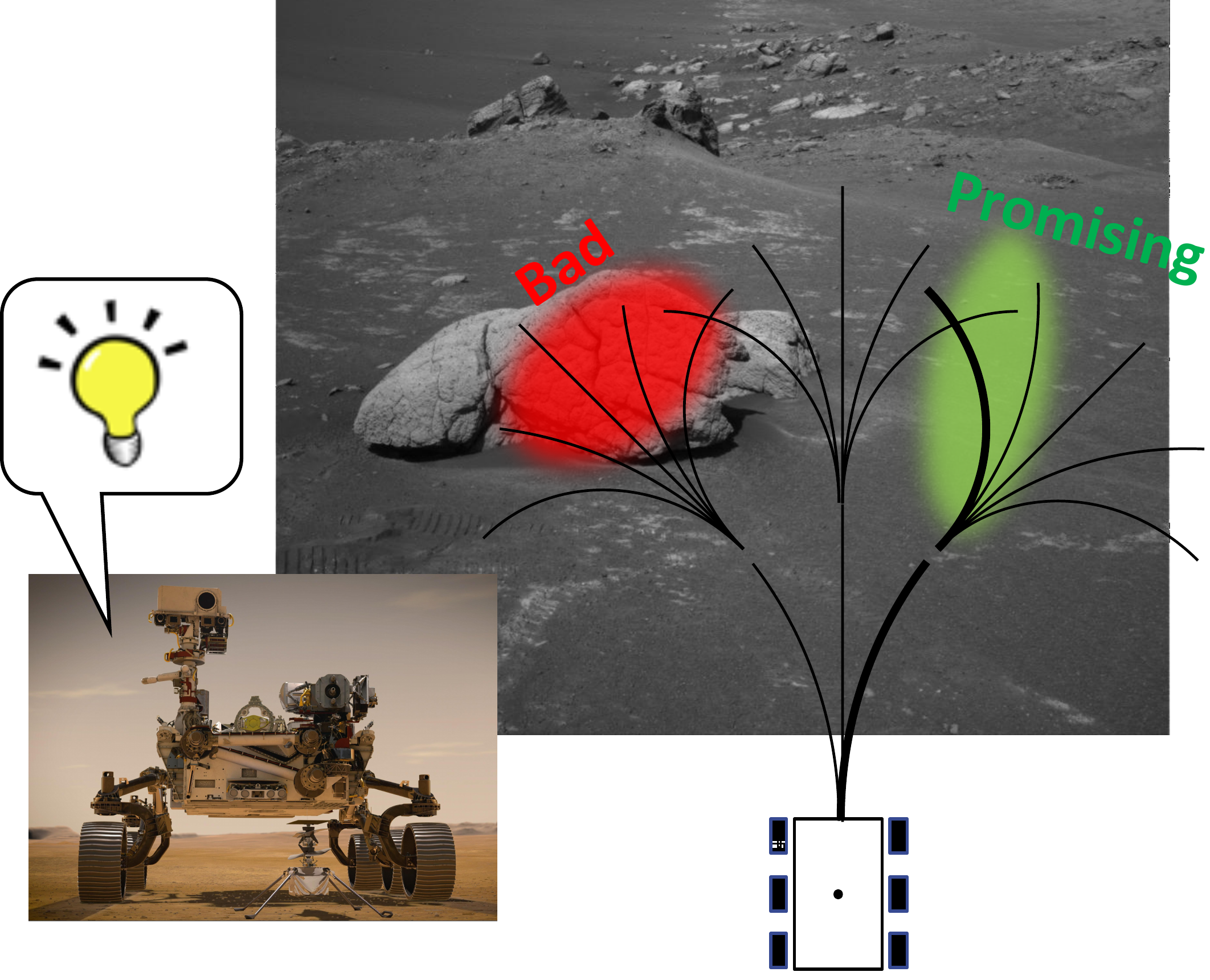}\\
    \vspace{-4mm}
    \caption{Experienced human rover drivers can intuitively find safe paths. The proposed learning-based heuristics is similar to human's intuition. It predicts which paths are likely feasible, based on a model learned from past examples.} % Image: NASA/JPL.}
    \vspace{-4mm}
    \label{fig:teaser}
\end{figure}

A key motivation for our approach is to study how best to effectively integrate learning-based methods with model-based planning frameworks, by carefully breaking down the traditional navigation stack, identify key components that are bottlenecked and integrate ML \emph{only} where it is most useful. 
While there has been a recent trend towards integrating learning into path planning and navigation \cite{xiao2020motion}, %have tried to alleviate some of the above mentioned limitations of classical planners. 
the use of learning methods also presents key challenges for real-world, safety-critical planning problems. First, learning-based methods are inherently unpredictable \cite{dhiman2018critical} and lack the necessary mathematical framework to provide guarantees on correctness, a strict necessity for path planning of safety-critical systems like our Mars rovers \cite{hunt2020verifiably}. Second, they are sample-inefficient to train, sub-optimal for goal-directed navigation tasks, and have so far been validated only in simple domains. Finally, they cannot exploit the compute vs performance trade-off of fine-grained planning to improve the real-time computational efficiency of the system. 
%Here, we argue that there is a growing need to effectively integrate learning-based methods with model-based planning frameworks by carefully breaking down the traditional navigation stack, identify key components that are bottlenecked and integrate ML \emph{only} where it is most useful. 

In this paper, we present \textit{MLNav} -- a learning-enhanced framework for real-time motion planning that takes the best of the two worlds: the ability to exploit the contextual structure of the environment from learning-based methods \textit{and} the ability of guaranteeing predictable and safe behaviors from classic search methods. In a nutshell, MLNav is a  search-based path planner that uses learned heuristics, where the safety of the chosen path is guaranteed by running a model-based collision checker.  The use of learned heuristics within search-based robotic planning has previously been studied \cite{smith2018learning,riviere2021neural,qureshi2020motion,choudhury2018data,ichter2018learning}, including showing favorable comparisons versus hand-craft heuristics for rover path planning \cite{abcouwer2020mlnav}.
%This work is motivated by our previous work \cite{abcouwer2020mlnav} that compared hand-crafted and learning-based heuristics, and concluded that the latter outperforms the former for Mars rover path planning. 
Our main contribution is proposing a general system design principle for effectively integrating ML-based approaches into existing navigation pipelines of safety-critical robotic systems, as well as a concrete instantiation for Mars rover navigation. 

Our system design is driven by a key observation that, in many safety-critical real-world planning setting, collision checking accounts for vast majority of the total computation time \cite{elbanhawi2014sampling}. 
This is because collision checking often involves model-based computation of the 3D geometry of all objects in the scene; 
furthermore, it has to run numerous times on path/motion options until finding the optimal (or at least feasible) one, particularly in a complex environment. This prompts the question: can we learn a proxy collision heuristic based on a planner's past experience? This learned heuristics, used as a subroutine, would then be able to guide the motion planner to select paths that have a high probability of being collision-free. In this way, the actual collision checks only need to be performed once on the final selected path for ensuring safety, saving significant computation.
%Second, the subroutine employed to compute the heuristic should not end up needing more computation than the original problem of collision checking itself. Using this knowledge, the question is how can we design our framework to exploit model architectures that are computationally efficient, sample-efficient to train, and need only a single forward inference per planning cycle. 

%% file: sec_contribution.tex
Our contributions are as follows:
\begin{itemize}
    %\item We build upon our previous work \cite{abcouwer2020mlnav} to propose a holistic framework for high-stakes planning that can effectively reason in complex environments while guaranteeing safety. Our approach makes judicious use of machine learning that enhances planning effectiveness (while guaranteeing safety) in a way that reduces computational and memory overhead. (Section \ref{sec:approach})
    \item We propose a holistic framework for high-stakes planning in complex environments, that makes judicious use of machine learning to enhance planning effectiveness (while guaranteeing safety) in a way that reduces computational and memory overhead. (Sec. \ref{sec:approach})
    \item We ground our framework in a real-(out-of-this-)-world path planning algorithm, ENav, where the goal is to efficiently find collision-free paths in hazardous terrains. We identify key components in the planning framework that are bottle-necked by poorly-informed reasoning and can benefit from using machine learning (Sec. \ref{sec:enav}).
    \item We evaluate using the flight software simulator (ENav Sim).  We use both a realistic environment model that was used for the development and testing of ENav for the Perseverance Rover (Sec. \ref{sec:exp}), as well as a real Martian terrain data collected by the Rover (Sec. \ref{sec:mars_exp}).
    \item Our experiments on real Mars data demonstrate a significant improvement in performance over the baseline (ENav), particularly in the number of collision check to run until finding a feasible solution, without any compromise in safety. In challenging cases, ENav fails to return a feasible path whereas MLNav succeeds.
\end{itemize}

%While we restrict our scope to mobile robots, the ideas discussed here are broadly applicable to other domains where real-time, safety-critical motion planning is used. 

%% Old version
% Our contributions are threefold:
% \begin{itemize}
%     \item Taking a system’s view of the motion planning pipeline for mobile robot navigation, we propose sensible modifications to bridge the gap between classical and learning-based methods, such that the resulting planner leverages advantages of learning where learning excels, and relies on model-based strategies where they excel.
%     \item We propose a novel formulation to learn a proxy collision map that effectively acts as a data-driven heuristic to guide trajectory search for fine-grained planning, while being allowing the planner to be both safe and computationally efficient by design. 
%     \item We evaluate our method beyond toy-datasets using a real-(out-of-this-)-world case-study of Mars Rover navigation, demonstrate the efficacy on challenging, high-fidelity simulated environments, and validate our representation through extensive experiments and ablation studies.
% \end{itemize}
% While we restrict our scope to mobile robots, the ideas discussed here are broadly applicable to other domains where real-time, safety-critical motion planning is used. 

%% file: sec_related.tex
Planning and navigation have a rich and varied history. In this review of related work, we focus primarily on contemporary machine-learning-based approaches.

\subsection{ML in Planning and Navigation}
The success of deep learning has made the use of ML in planning approaches attractive, as it opens up the possibility of learning high-capacity models to reason about complex, real-world environments. Several prior work proposed systems that use of end-to-end navigation approaches that attempt to either replace the entire navigation pipeline from perception to control using black-box image-to-action mappings \cite{levine2016end,daftry2016learning,pfeiffer2017perception} or attempt to learn an end-to-end cost mapping based on large amounts of expert demonstrations \cite{abbeel2004apprenticeship,ziebart2008maximum,wulfmeier2017large,wulfmeier2016watch}. While considerable progress is being made in this direction, they are several challenges towards adoption in real-world safety-critical robotic applications like Mars rovers. In contrast, a few have attempted to replace only particular navigation subsystems such as global planning \cite{yao2019following}, local planning \cite{gao2017intention,faust2018prm} or improve individual components such as world representation \cite{richter2017safe}. Our work falls in this latter category, where we try to systematically “unwrap” our Mars rover system and identify collision checking as a bottleneck component that can be improved using ML. We also refer the readers to \cite{xiao2020motion} for a comprehensive review on this topic. 

\subsection{ML for Collision Checking}
Several related work have also tried to use machine learning to overcome the computational bottleneck of collision-checking. Examples include learning a distribution of promising regions \cite{ichter2018learning}, learning heuristics for collision distance metrics such as swept volume \cite{chiang2018fast}, employing a lazy approach to evaluate only the most promising edges based on predicted energy costs  \cite{choudhury2018data}, and learning to accelerate collision checking itself by modeling the configuration space of the robot \cite{das2020learning,han2020configuration,huh2016learning,kew2019neural}. % 
While similar in theme, in the sense they all use learning to improve the bottlenecks of collision checking, all the above approaches, vary considerably in system design.
The key design choices are: (1) to operate directly from raw sensing data or a processed representation (e.g., a configuration space); (2) to directly predict the probability of collision or some other measure such as which paths are more ``promising''; and (3) the interplay with the model-based collision checker and overall system which ultimately guarantees safety.  Which design choices work best (including safety guarantees, and computational efficiency) depends on the system and its goals. To the best of our knowledge, our approach is the first to be validated on a mature system with real-world safety-critical needs.

%While similar in theme, in the sense they all use learning to improve the bottlenecks of collision checking, all of the above methods are not directly compatible for use on the Mars rover navigation system. 

%% file: sec_approach.tex
\subsection{Overview and Problem Formulation}
\label{subsec:problem}
\begin{figure}[t]
    \centering
    \includegraphics[width=0.9\linewidth]{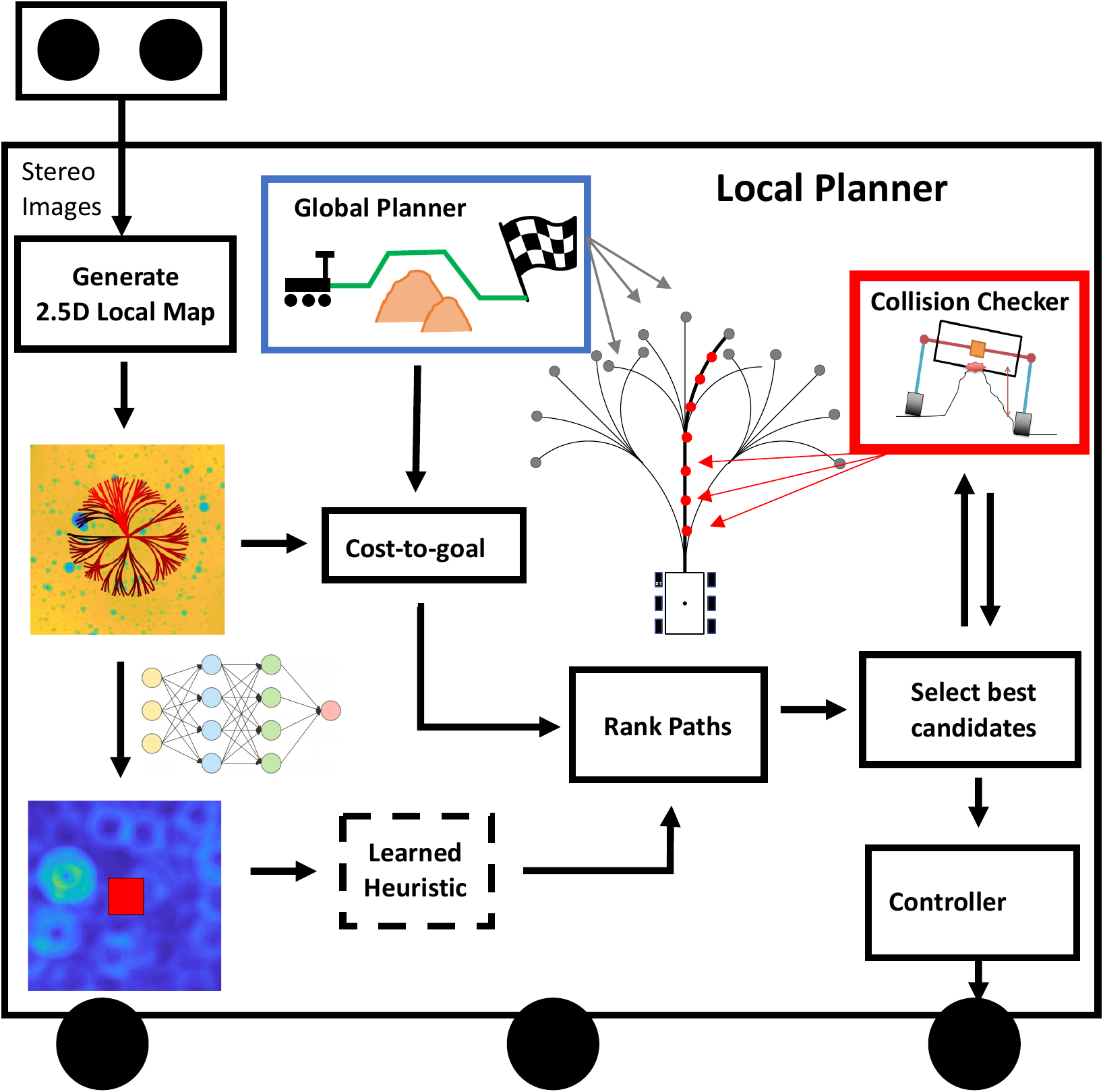}
    \caption{MLNav Framework: A classical search-based planner is augmented with a learning-based heuristic to accelerate the search, where the safety of the selected path is guaranteed by a model-based collision checker.}
    \label{fig:mlnav-overview}
    \vspace{-4mm}
\end{figure}

In this section, we provide a general formulation of our proposed MLNav framework (as shown in Figure \ref{fig:mlnav-overview}). Traditionally, the motion planning pipeline for goal-directed real-time autonomous navigation is done in hierarchical receding horizon manner, with a global planner at the top level driving the robot in a general direction of the goal while a local planner uses the immediate perceived environment (up to a finite sensing horizon) to makes sure that the robot avoids obstacles while making progress towards the goal. 

For dynamically constrained systems operating in high-dimensional complex environments, a library of candidate trajectories is usually computed offline by sampling from a much larger (possibly infinite) set of feasible trajectories. Such libraries effectively discretize a large control space and enable tasks to be completed with reasonable performance while still respecting computational constraints. Such library-based model predictive approaches have been widely used in state-of-the-art robotic systems \cite{dey2016vision}. We follow a similar paradigm as well for rovers on Mars \cite{carsten2007global}, and use it for our MLNav framework. Note that the proposed architecture does also generalize to any sampling-based motion planning problem where edge-evaluation is expensive.
%leveraged by most DARPA challenges \cite{urmson2006robust,urmson2008autonomous}, autonomous ground \cite{richter2014high} and aerial vehicles \cite{dey2016vision,choudhury2014planner}, as well as rovers on Mars \cite{carsten2007global}. Here, we follow a similar paradigm to formulate our problem. 
Let us define a robot at time $t$ with state $\phi(x_t, m)$, where $x_t \in R$ is the pose of the rover operating in a 2.5D static local heightmap $m \in \mathcal{M}$, sampled from a distribution of terrains $p(m)$. Also, let us assume a library $\mathcal{L}$ of $N$ trajectories is given, such that $\mathcal{L}$ = \{$\xi_j$\}, j = 1: N, $\xi_j \in \Xi$, where $\Xi$ spans the space of all possible trajectories. At each planning cycle the local planner picks the trajectory that yields the least cost $C(\cdot)$ for traversal. We formulate this as an optimization problem:
\begin{equation}
    \phi_{o} \mapsto \overset{*}{\xi} = \argmin_{\xi  \in \mathcal{L}} C(\xi_j)
\label{eq:pathplanning}
\end{equation}
\noindent
where $\phi_{o}$ is represents the initial state of the robot in the map. The cost function being optimized is then defined as:
\begin{equation}
C(\xi_j) = \alpha \cdot C_{goal}(\xi_j, \phi_{o}) +  \beta \cdot \sum_{t = 1}^{T} C_{collision}(\xi_j, \phi(x_t, m)))
\label{eq:cost}
\end{equation}
\noindent
where $C_{goal}$ is the cost for path execution, such as the time to get to the final goal, consisting of the cost within the planning horizon and the cost-to-go beyond the horizon, which usually comes from a separate global planner. This part is fast to compute but does not account for vehicle safety. $C_{collision}$ is the expected collision cost for each trajectory, computed by estimating the clearance of the robot with the local terrain features over the planning horizon of $T$ time steps. In a deterministic environment $C_{collision} \in \{0, \infty\}$; in an uncertain environment, it represents the probability of collision.
The collision cost is typically computed by repeatedly running a collision checking algorithm at a certain interval over the candidate trajectories.
The computation time of collision cost grows proportionally with both the number of trajectory options $N$ and the length of the horizon $T$.

\subsection{Learned Proxy Collision Heuristics}
We propose to alleviate this bottleneck by leveraging the planner's past experience to learn a proxy collision heuristic: 
\begin{equation}
    C_{proxy\_collision}(\mathcal{L}) = f_h(\phi(x, m))
\end{equation}
where the function $f_h$ maps the rover's state and local terrain features, $\phi(x, m)$, to a probability of collision. While any ML model can be used to learn $f_h$, we propose to use a Convolutional Neural Network (CNN) based image-to-image translation model. This has two key advantages: First, since the heightmap $m$ is usually encoded as an image, it allows us to estimate $f_h$ using a single-shot inference on the entire map. The subroutine that estimates this heuristic could, in principle, be queried each time we want to estimate $C_{collision}$ in Equation \ref{eq:cost} using only local features and robot pose. However, that would be helpful only if we the cost of invoking $f_h$ each time is significantly lower than the cost of computing the actual collision cost. Second, it allows us to leverage the representation power of CNNs without handcrafted feature engineering. Once the heuristic map is estimated, computing the $C_{collision}(\xi_j, \phi(x_t, m))$ values in Equation \ref{eq:cost} gets reduced to a trivial constant-time look-up.

The proxy collision heuristic is then used to select the optimal trajectory $\overset{*}{\xi}$ for execution. Importantly, the model-based collision checking algorithm is then run only on the optimal path. If it is not feasible, it evaluates the next best path until a feasible one is found.  
In an ideal case where $f_h$ makes a perfect prediction of $C_{collision}$, the planner finds the optimal path by running collision checking only on a single path. Regardless of the performance of $f_h$, the safety of the chosen path is always guaranteed because collision checking must pass for a path to be executed. This allows us to leverage the benefits of ML based models while maintaining the same safety guarantees of model-based planners.

%% file: sec_enav.tex
% \begin{figure*}[!t]
%     \centering
%     \includegraphics[width=\linewidth]{figures/enav-overview-v2.pdf}\\
%     \caption{Overview of ENav and Simulator used for data collection}
%     \label{fig:ml-ace}
% \end{figure*}

Next, we ground our MLNav framework in a specific problem domain and study it in the context of path planning for Mars rovers, based on the ENav algorithm. 
%Rover navigation has been traditionally limited by the available computational power in space; when driving autonomously, the limited computation means that the rover must efficiently identify a hazard-free path. Failing to do so will result in restricted driving duty cycle that reduces the rover’s average traverse rate. Future missions with significantly increased requirements on driving distance demand more agile and autonomous mobility. Although the autonomous navigation software is usually able to guide the rover to the goal, there are known circumstances where it is susceptible to failure, particularly in a highly cluttered environment. For example, Monte Carlo simulations show that ENav would fail to complete a 80m drive for $10\%$ of the time on Jezero Crater. These failures are largely attributed to the planning approaches in ENav. Here, we consider the ENav algorithm as our baseline and starting point, and integrate the MLNav extensions.

\subsection{Overview of ENav}
ENav is a tree-based planner that considers a parametrized tree of candidate trajectories, represented by $\mathcal{L}$ in Eq. \ref{eq:pathplanning}. At each planning cycle, it generates 
a 2.5D terrain height map $m$ from stereo imagery.
$C_{goal}$ is the estimated time to reach the final goal, with an additional penalty based on terrain roughness. 
$C_{collision}$ is computed by a collision checking algorithm called Approximate Clearance Evaluation (ACE) \cite{otsu2020fast}. 
The computation of $C_{collision}$ is by far more complex in comparison to $C_{goal}$. Therefore, to save the precious on-board computational resource, ENav gives up finding the optimal path and instead makes the following modifications to optimal path planning in Eq. \ref{eq:pathplanning}. 
First, before running ACE, ENav computes $C_{goal}$ for all candidate trajectories in $\mathcal{L}$ and sort them by $C_{goal}$.
Then, ENav greedily evaluates $C_{collision}$ of the trajectories from the top of the sorted list.
It cuts off the search if at least one feasible path is found \textit{and} a pre-specified threshold on the number of ACE execution is reached, even if some trajectories in $\mathcal{L}$ remain unevaluated. 
Finally, ENav chooses the best trajectory among the ones that have evaluated before the cut off. 

The vast majority of the computation time of ENav is for computing $C_{collision}$ with ACE. 
In the worst case where a sole feasible path is ranked at the bottom of the sorted list, ENav needs to run ACE on all candidates in $\mathcal{L}$;
if there are multiple feasible paths but the optimal one is located below the cut off threshold, a suboptimal path may be chosen. 
This is why ENav performance could be enhanced by introducing a heuristic to better sort $\mathcal{L}$.  

%The sorted list of paths are then greedily evaluated using a safety check algorithm known as Approximate Clearance Evaluation (ACE) \cite{otsu2020fast}, starting with the lowest cost path. ACE obtains conservative bounds on vehicle clearance, attitude, and suspension angles by estimating the lowest and highest heights that each wheel may reach given the underlying terrain, and calculating the worst-case vehicle configuration associated with those extreme wheel heights. The bounds are guaranteed to be conservative, hence ensuring vehicle safety during autonomous navigation. If the bounds violate parameterized limits, ACE can return an infinite cost, rendering that path infeasible. Otherwise, a finite cost is returned based on proximity to the limits, which accumulates over the length of the paths. Note that ACE cost here is equivalent to the $C_{collision}$ in the general formulation.

%Once a path with finite cost is found, ENav continues to analyze other candidate paths until reaching a threshold, and returns the lowest-cost ACE-validated path found for execution. ACE is pivotal for maintaining the safety of the rover, but running ACE on many poses along many potential paths represents a significant computational burden. This is most evident when the initial sorting of the paths is poor with respect to terrain safety and the most promising paths all fail ACE safety evaluations, which is more likely in complex and challenging terrain. 

\subsection{MLNav implementation on ENav}
In order to more effectively sort ENav's list of candidate rover paths, such that the most highly ranked paths are feasible and near-optimal, we use the learned proxy collision heuristics. The updated cost function then is as follows:
\begin{equation}
    C_{total}(\xi_j) = \alpha \cdot C_{goal}(\xi_j, \phi_{o}) +  \beta \cdot C_{proxy\_ACE}(\mathcal{L})
\label{eq:enav-cost}
\end{equation}
where $C_{proxy\_ACE}$ is computed as a look-up using the predicted heuristic map. %The total cost $C_{total}$ is then both goal- and collision-aware. 

\begin{figure}[t]
    \centering
    \includegraphics[width=\linewidth]{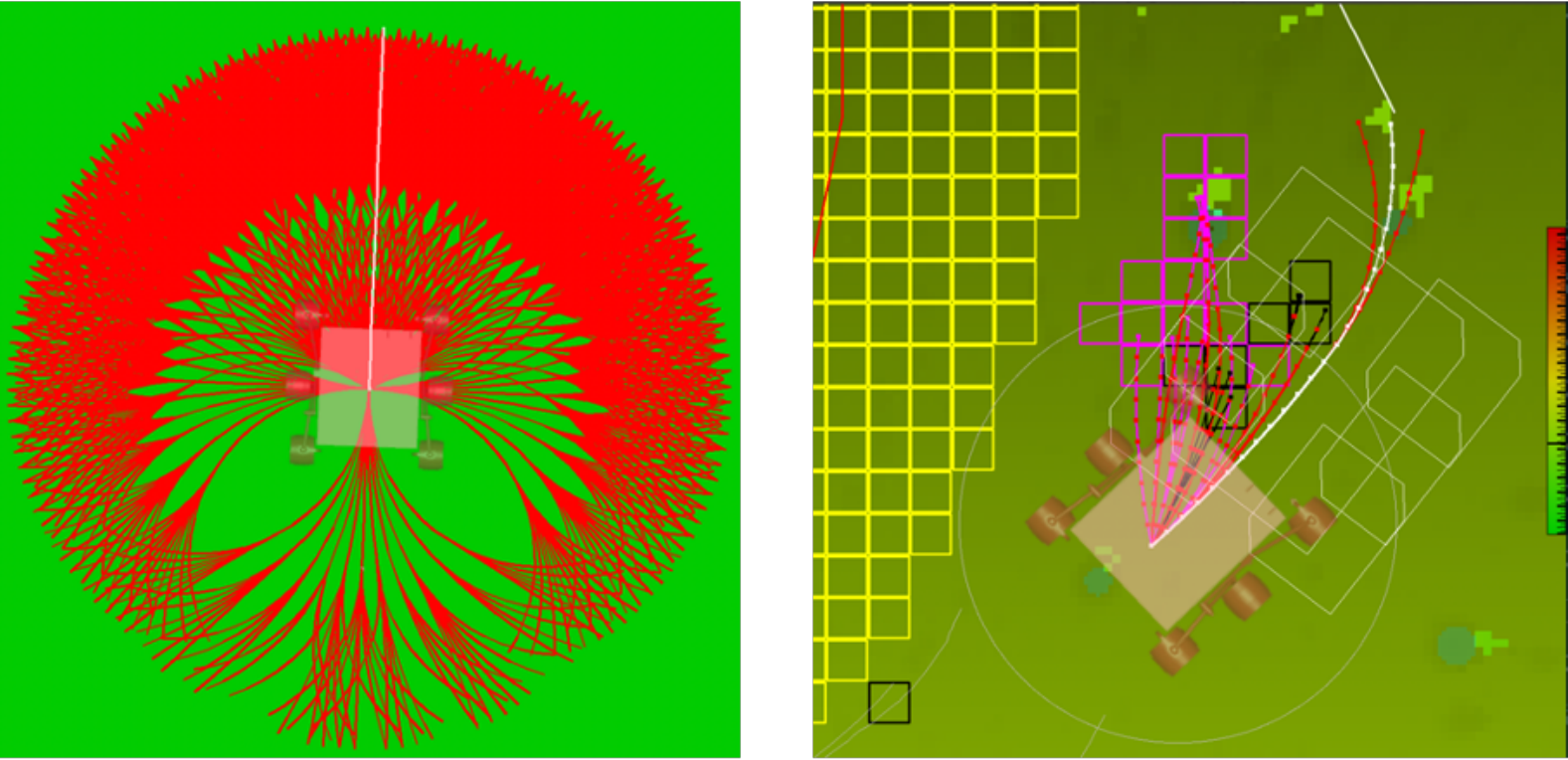}\\
    \vspace{2mm}
    \includegraphics[width=\linewidth]{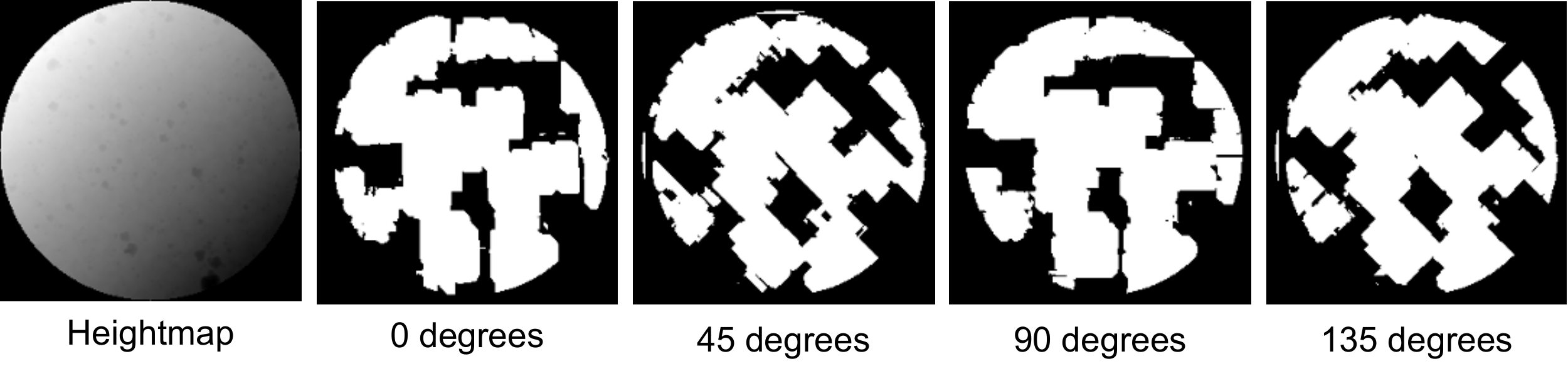}
    \vspace{-7mm}
    \caption{ENav overview and examples of training data for learning the heuristic function. For each heighmap, a set of 8 output ACE Maps were generated, corresponding to different rover headings.}
    \label{fig:train-data}
\end{figure}

In our particular implementation for MLNav, we use a model based on a U-Net \cite{ronneberger2015u}, and learn it in a supervised manner. The input is a height-map $m$ and the output is a proxy collision heuristic map, such that the value for each pixel in the heuristic map corresponds to the predicted collision probability. The collision cost at any given spatial point in the height-map depends not only on the terrain features but also on the rover heading. We encode the rover heading as part of the learning problem itself by extending the model output to have a multi-channel representation such that each channel represents a cardinal heading angle (see Figure \ref{fig:train-data}). Note that the granularity of the discretization depends on the specific instantiation of the MLNav framework. For this implementation, we found a discretization of 8 heading angles (at 45 degree intervals) to be sufficient. Sigmoid activation is then applied to each channel to give a value in the range [0, 1]. Training data is collected by running ACE on synthetic terrains that are representative of Martian terrain. Note that the training set does \textit{not} include real Mars terrains although our test does, as described in Section \ref{sec:mars_exp}.

%, assuming that running offline execution of expensive collision checks and Monte Carlo trials is cheap. 

%% file: sec_experiments.tex
\subsection{Experimental Setup}
For all our experiments presented in this section, we used a ROS-based, high-fidelity simulation environment called ENav Sim \cite{toupet2020ros} that was originally developed for prototyping and testing of Perseverance's ENav algorithm. ENav Sim is capable of generating a rich set of varied 2.5D terrains representative of the candidate Mars landing sites, producing synthetic stereo images for simulating onboard hightmap generation, and simulating the rover's motion. At its core, the simulator wraps a flight software implementation of ENav and a software library called HyperDrive Sim (HDSim), which has been used for terrain simulation for multiple Mars rover missions. HDSim provides an image rendering capability based on the rover's navigation cameras, rover-terrain settling and a realistic slip model. Furthermore, ENav Sim provides a method for running a large-scale Monte Carlo simulation in parallel and automatically generate reports that capture the key ENav performance metrics. The learned heuristic model was implemented in Tensorflow and setup as a separate subroutine, that could be invoked by the ENav algorithm on-demand as a ROS service call.

%%***********************************************************************************
%Performance Evaluation of MLNav for Mars Rover Navigation. We compare MLNav to the baseline ENav, and also characterize the sensitivity to key parameters and tree design. Baseline$^{\dagger}$ and MLNav$^{\dagger}$ represent results from trials where the minimum number of ACE threshold was set to 0. MLNav$^{\dagger} (BT)$, MLNav$^{\dagger} (DT)$ and MLNav$^{\dagger}$ represent result from trials where the tree design has a greater branching factor, a deeper tree and a complex tree with variable length arcs. See \ref{sec:exp-sim}(C) for more details.

\begin{table*}[!t]
    \caption{Performance Evaluation of MLNav for Mars Rover Navigation. We compare MLNav to the baseline ENav, and study its sensitivity to key parameters (MLNav$^{\dagger}$) and tree design (MLNav$^{\dagger} (BT)$, MLNav$^{\dagger} (DT)$ and MLNav$^{\dagger}$). See \ref{sec:exp-sim}(C) for more details.}
    \label{tab:enav-ablation}
    \centering
    \begin{tabular}{c c c c c c c c c}
    \toprule
    \\[-2mm]
    \textbf{Metric} & \textbf{Terrain}  & \textbf{Baseline} & \textbf{MLNav} & \textbf{Baseline$^{\dagger}$} & \textbf{MLNav$^{\dagger}$} & \textbf{MLNav$^{\dagger}$ (BT)} & \textbf{MLNav$^{\dagger}$ (DT)} & \textbf{MLNav$^{\dagger}$ (VLT)}\\
    \\[-2mm]
    \midrule
    \midrule
    \multirow{2}{*}{Success rate (\%)} & Benign & 100.0 & 99.5 & 99.4 & 99.9 & 99.9 & 98.5 & 99.7\\
                                       & Complex & 69.9 & 72.5 & 69.0 & 69.3 & 73.9 & 59.9 & 78.8\\
    \midrule
    \multirow{2}{*}{Path Inefficiency (\%)} & Benign & 4.4 & 3.9 & 3.95 & 3.3 & 3.2 & 3.7 & 3.0\\
                                       & Complex & 25.4 & 20.4 & 22.1 & 19.5 & 17.6 & 19.1 & 17.6\\
    \midrule
    \multirow{2}{*}{Number of Collision Checks} & Benign & 275 & 262 & 74 & 39 & 58 & 78 & 70\\
                                       & Complex & 377 & 283 & 216 & 90 & 142 & 164 & 317\\
    \midrule
    \multirow{2}{*}{Overthink Rate (\%)} & Benign & 5.3 & 2.2 & 4.7 & 1.2 & 2.7 & 4.4 & 2.98\\
                                       & Complex & 20 & 7.1 & 19 & 6.5 & 10.3 & 15.6 & 12.6\\
    \bottomrule
    \end{tabular}
    
\end{table*}

\begin{table}[t]
    \centering
    \caption{Performance Evaluation vs Model Size}
    \begin{tabular}{c c c c c}
    \toprule
    \\ [-2.5mm]
    \textbf{Models} & U-Net & SegNet & DeeplabV3+ & PSPNet \\
    \\ [-2.5mm]
    \midrule
    \\ [-2.5mm]
    Model Size & 130 MB & 110 MB & 190MB & 280MB \\
    \\ [-2.5mm]
    Accuracy & 95.1\% & 81.7\% & 82.8\% & 85.2\% \\ 
    \\ [-2.5mm]
    \bottomrule
    \end{tabular}
    \label{tab:mlace-models}
\end{table}

\subsection{Training and Validation of Learned Heuristics}
We first evaluated the fidelity of our proposed model to learn the proxy collision heuristics. Training data was gathered by running Monte Carlo simulations of the baseline ENav algorithm on 1500 terrains and randomly sampling 8 heightmaps from each trial. For each cell in each sampled height-map, the ACE costs were estimated for eight fixed rover heading values at $45\degree$ intervals, resulting in a 8-channel “ACE map” such that each channel corresponded eight heading-specific ACE values. This ACE-map was then used as the training signal for our modified U-Net model. A total of 12000 height-map and ACE-map pairs were generated; 9500 were used for training, and 2500 were used for validation. The learned heuristic model was able to achieve $97.8\%$ training accuracy and $95.1\%$ validation accuracy, demonstrating that the model was able to accurately learn the mapping to collision probabilities directly.

% \noindent
% \textbf{Ablation Studies:}
Next, we performed an ablation study to justify the design choice towards using the U-Net model architecture. In particular, the MLNav framework was designed with computational efficiency as the primary focus. However, one might ask the question: does using a larger state-of-the-art models from semantic segmentation help? Table \ref{tab:mlace-models} shows the performance of different models on learning the proxy collision heuristics. We observe that U-Net provides the best performance even though it is the smallest model. Our hypothesis is that while bigger models have a much larger capacity to learn generic representations required for visual recognition tasks, they also require larger training sets. In contrast, U-Net was specifically designed for biomedical applications, where it could be trained with very few images. 

\subsection{Benefits to Mars Rover Navigation}
\label{sec:exp-sim}
In this section, we evaluate the potential benefits to the overall pipeline, in the context of the rover navigation. Monte Carlo simulations were run for both the baseline ENav and MLNav, using the same set of terrains. Note that the terrains used for these experiments are a separate set from the set of terrains used to train the learned model, to ensure that the observed performance is not biased. The Monte Carlo simulation consisted of 1500 terrains with various slope and rock density, which is quantified by CFA (cumulative fraction of area) \cite{golombek2003rock}. We report out results on two categories: (1) Benign - terrains with slope less than $15^{\circ}$ and CFA value of $7\%$ or less, and (2) Complex - terrains with greater slope or CFA. For ENav, the default tree of paths used in our experiments is composed of 14 candidate turns-in-place, followed by 11 3-meter arcs of various curvatures, followed by another set of 11 3-meter arcs. Thus, at each step the rover is planning about 6 meters ahead of its current position, and considering 1694 potential paths. Note that the hardware of the Perseverance rover does not allow steering while driving, and thus the rover can only move in fixed-curvature arcs.
%While both categories are important, our primary goal is to improve performance on complex terrains. 
For evaluation, the following performance metrics were used: 
\begin{itemize}
    \item \textbf{Success Rate:} Defined as the percentage of trials that result in the rover reaching the goal without timing out or violating safety constraints. Higher values are better.
    \item \textbf{Path Inefficiency:} Defined as the average excess length of the path taken by the rover, as compared to a straight line path from straight to goal, expressed as a percentage. For example, if the rover drives 110 m to reach a goal that is 100 m in straight line distance, the path inefficiency is 10 \%. Lower values are better.
    \item \textbf{Number of Collision Checks:} Defined as the average ACE runs per planning cycle.
    %Using knowledge of the actual computation time of each ACE check on a target hardware, a corresponding cycle time can be calculated. 
    Lower values are better.
    \item \textbf{Overthink Rate:} Defines as the average number of planning cycles required above a minimum number of ACE checks (default: 275), defined in percentage. When the number of ACE checks exceed a threshold, it indicates that the highest ranked paths were all deemed unsafe, and the rover may need to stop and ``overthink" until a solution is found. Lower values are better. 
\end{itemize}

Table \ref{tab:enav-ablation} shows the results from the MC simulations. We observe that as using the learned heuristics there is a significant improvement in the overall efficiency of the ENav algorithm. In particular for complex terrain, there was a $20\%$ reduction in path inefficiency, $25\%$ reduction in number of collision checks and a $65\%$ reduction in overthink rate using the MLNav framework, as compared to the baseline ENav algorithm. This improvement in computational and performance efficiency comes at almost no cost; the success rate of MLNav is comparable to the baseline, and within the experimental margin of error. Furthermore, one of the primary objectives of this work was to design a framework that could leverage the benefits of ML without compromising on safety guarantees of the vehicle. Based on our extensive MC simulations, our assertion holds. In all our experiments, no trial failures were reported due to violated safety constraints; all failures are due to either timeouts or failures to find paths to the goal. This is the most significant result of this paper.

\subsection{Sensitivity \& Ablation Analyses}
Next, we performed ablation studies on the sensitivity of the MLNav framework to two key parameters:
\subsubsection{Performance vs. minimum ACE threshold} For all our experimental trials above, we define a threshold value of 275 for the minimum number of ACE checks. In practice, this number can be thought of as the computational budget allocated to the planner and also protects against choosing high-risk paths with finite but poor ACE cost. However, if the learned heuristics is working optimally, one could imagine that further evaluation beyond the first safe trajectory is no longer needed and this computation might be better used elsewhere. In order to study this trade-off we set the minimum ACE thresholds to 0 and repeated the above experiments. We call this version MLNav$^{\dagger}$, and tabulate the results in Table \ref{tab:enav-ablation}.
For comparison, we also run ENav with the minimum ACE threshold set to 0, shown as Baseline$^{\dagger}$ in the table.
This intuitively means that MLNav$^{\dagger}$ and Baseline$^{\dagger}$ always choose the first feasible path found and does not search any further.
Interestingly, the difference in path inefficiency between MLNav and MLNav$^{\dagger}$ (also between Baseline and Baseline$^{\dagger}$) is within the error margin. 
Furthermore, MLNav$^{\dagger}$  substantially improved on the number of collision checks and overthink rate. Since ACE is run every 25 cm, evaluating a single 6 m path requires 24 ACE runs. Therefore, the average number of ACE checks being 39 and 90 for benign and complex terrains respectively means that MLNav$^{\dagger}$ finds a feasible path only after evaluating 1.6 and 3.7 options. 
These results imply that the learned heuristics almost always ranks the optimal path near the top of the list. 

\subsubsection{Performance vs. Tree Design} The above experiments  demonstrated that MLNav can significantly reduce computation time in the ENav planning. We next ask: Can this surplus computation be put to use back in planner to improve the success rate? We evaluate three different approaches to increase the complexity of the tree. First, by increasing the number of candidate trajectories in the library, which can increase the probability of finding a safe and efficient path towards goal. Here, we increase the branching factor of possible actions at each tree-depth by $4$, leading to 4050 candidate trajectories. Note: this could increase the computation time by 2-3x for the baseline ENav algorithm. The results in Table \ref{tab:enav-ablation} - MLNav$^{\dagger}$(BT), shows a small improvement in overall success rate with further reduction in computation. Second, we use a deeper tree by adding a set of 11 arcs to previous leaf nodes, extending the planning horizon to 9m. This led to poor results (tabulated at MLNav$^{\dagger}$(DT)), where the success rate dropped to $59.9\%$ without any improvement in other metrics. We believe this is due to other system-level uncertainties such as stereo, slip, etc which may be significantly worse that far out. Finally, we use a more complex tree design with variable length arcs at different depth layers - the first layer was 11 turn-in-place, followed by 15 1-meter arcs, 11 2-meter arcs and 7 3 meter-arcs. The results using this tree design (tabulated as MLNav $^{\dagger}$ (VLT)) show an increase in success rate to $78.8\%$. As the average number of collision checks for MLNav $^{\dagger}$ was still lower compared to the baseline, we can conclude that with MLNav, now we can indeed use a more complex tree design to achieve a higher success rate in complex terrains without increasing the computational budget.

% \subsubsection{Performance vs. Cost Weights} Here we study the impact of $\alpha$ and $\beta$ parameters in Equation \ref{eq:enav-cost}. In particular these parameters control how aggressively we use the learned heuristic as compared to the $C_{goal}$, and captures the system's fidelity to trade-off between safety and optimality. For simplicity, we keep the $\alpha$ parameter constant, and only vary $\beta$. Note that $\beta = 0$ implies we choose paths only based on $C_{goal}$. As shown in \ref{fig:enav-weight}, with higher $\beta$ values (weighing the learned heuristic more aggressively) improves performance across all the metrics except success rate, as expected. In particular, the overthink rate decreases by almost $70\%$ with an aggressive reliance on the learned heuristic.

% \begin{figure}[!t]
%     \centering
%     \includegraphics[width=\linewidth]{figures/mlace-beta-combined.pdf}\\
%     \caption{Performance Evaluation vs. Cost Weights}
%     \label{fig:enav-weight}
% \end{figure}

% %%***********************************************************************************
\subsection{Hardware-in-the-loop (HIL) Benchmarks}
\label{sec:exp-hils}
%So far we have reported any gains in performance only in relative terms as the average reduction in number of collision checks (or ACE evaluations). 
We first benchmark using the RAD750 processor, which is similar to the one running on the Perseverance Rover. We find that it takes 12ms, on average, for each ACE check. For the baseline ENav algorithm this would translate to a total cost of ACE checking being 4.5s (377x12ms). In comparison, for the MLNav$^{\dagger}$ case, the total cost of ACE checking would be only 1.1s, saving 3.4s of computation time per planning cycle. %However, an important caveat is that if the time savings by using MLNav are completely consumed by the increased time used to calculate the heuristics, these claims are no longer valid. In theory, we have designed our framework with principles that would help with this but here we perform computational benchmarks in actual flight-like computing environments to settle this concern. 
To quantify the computational cost of running our learned model, we benchmarked its inference time using a Nvidia Jetson TX2, which serves as a reliable analog for future High-performance Space Computing (HPSC) \cite{doyle2013high}, and found that forward inference for predicting the collision heuristic map takes only 125ms, even without model optimization. We thus expect that a real-world deployment of MLNav will lead to significant and tangible acceleration of traditional navigation pipelines
%While we remark that this is not a fair comparison, as both the systems have not be tested on the same computing hardware, given the extremely with these numbers that actual real-world deployment of the learned heuristic model would lead to significant and tangible acceleration of traditional pipelines.

%% file: sec_mars_experiment.tex
\subsection{Experimental Setup}
%Experiments on Mars are out of the scope of this work because deploying a new algorithm like MLNav on a real Mars rover requires thorough verification and validation of safety requirements, as well as years of testing and integration with the full flight system. 

\begin{figure}[!t]
    \centering
    \includegraphics[width=\linewidth]{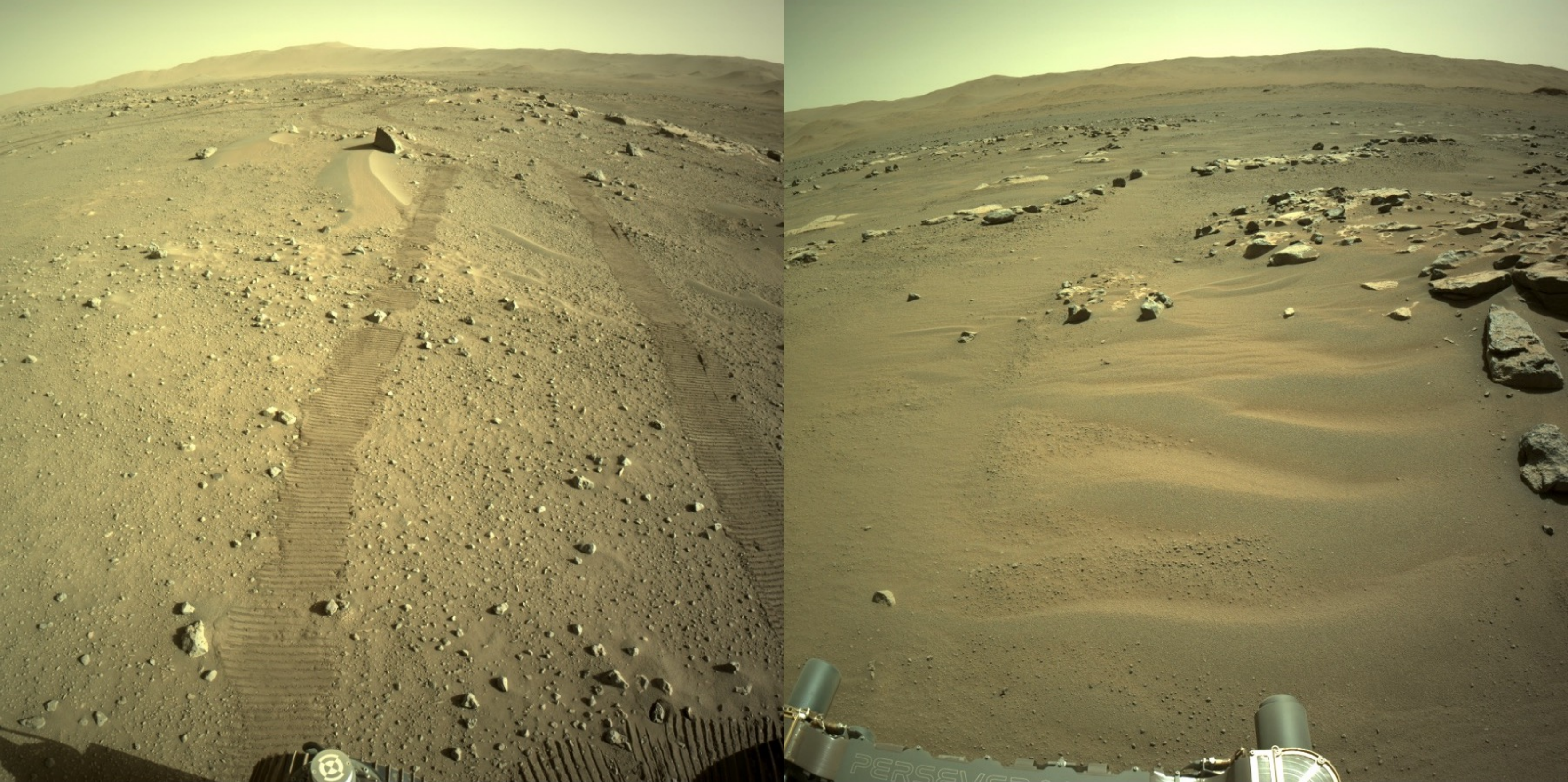}
    \vspace{-0.27in}
    \caption{Terrains used for experiments. Images taken by Perseverance on Sol 122 (left) and 178 (right). Image: NASA/JPL-Caltech}
    \label{fig:hafiq}
\end{figure}

We now validate MLNav on ENav Sim using real data from Mars collected by the Perseverance rover. A challenge to this approach is that we do not usually have the complete set of onboard images for reconstructing the terrain due to the limitation in communication data volume between the two planets. In particular, only the last and the penultimate images are typically transmitted for manual drives and, as of the writing of this paper, there have only been 17 occasions where the rover drove fully autonomously. Within these limitations, we chose two data sets acquired on the following Sols (Martian days since landing) for our test venues: 
\paragraph{Sol 122} The drive on Sol 122 was the last of a series of first-time activities (FTAs) for commissioning ENav on Perseverance. On this Sol, the rover was commanded to drive fully autonomously over $\sim$30 m northwards. As the ``final test" for ENav, the goal was intentionally set behind a rock, seen in Figure \ref{fig:hafiq}-left, such that the rover had to deviate from the straight-line path to get to the goal. Due to the need for detailed assessment of this particular drive, all of the on-board stereo image pairs were transmitted to Earth, which allowed us to reconstruct the terrain completely. ENav completed the drive successfully. 
\paragraph{Sol 178} It was one of the most challenging drives of Perseverance to date. The $\sim$85 m drive started with $\sim$15 m of manual driving, followed by fully autonomous driving along a ridge that concluded by climbing a slope to reach a science target. There were large rocks and exposed bedrock along the ridge and on each side of the slope as seen in Figure \ref{fig:hafiq}-right. Image pairs and onboard heightmaps were downlinked from the last $\sim15$ m segment of this drive, allowing us to recreate the corresponding portion of the terrain  in simulation. ENav completed the drive successfully.

For each Sol, we first ran stereo processing to produce DEMs (digital elevation models). The DEMs were then mosaiced using the onboard pose updates to reconstruct the 3D terrain with a 0.05 m resolution. In simulation, the start and goal were chosen to closely match Perseverance’s actual drive path. We used the MLNav$^{\dagger}$ (VLT) setting. Other parameters were identical to the ones we used in the actual Sol 122 and 178 drives.

\begin{figure}[!t]
    \centering
    \includegraphics[width=\linewidth]{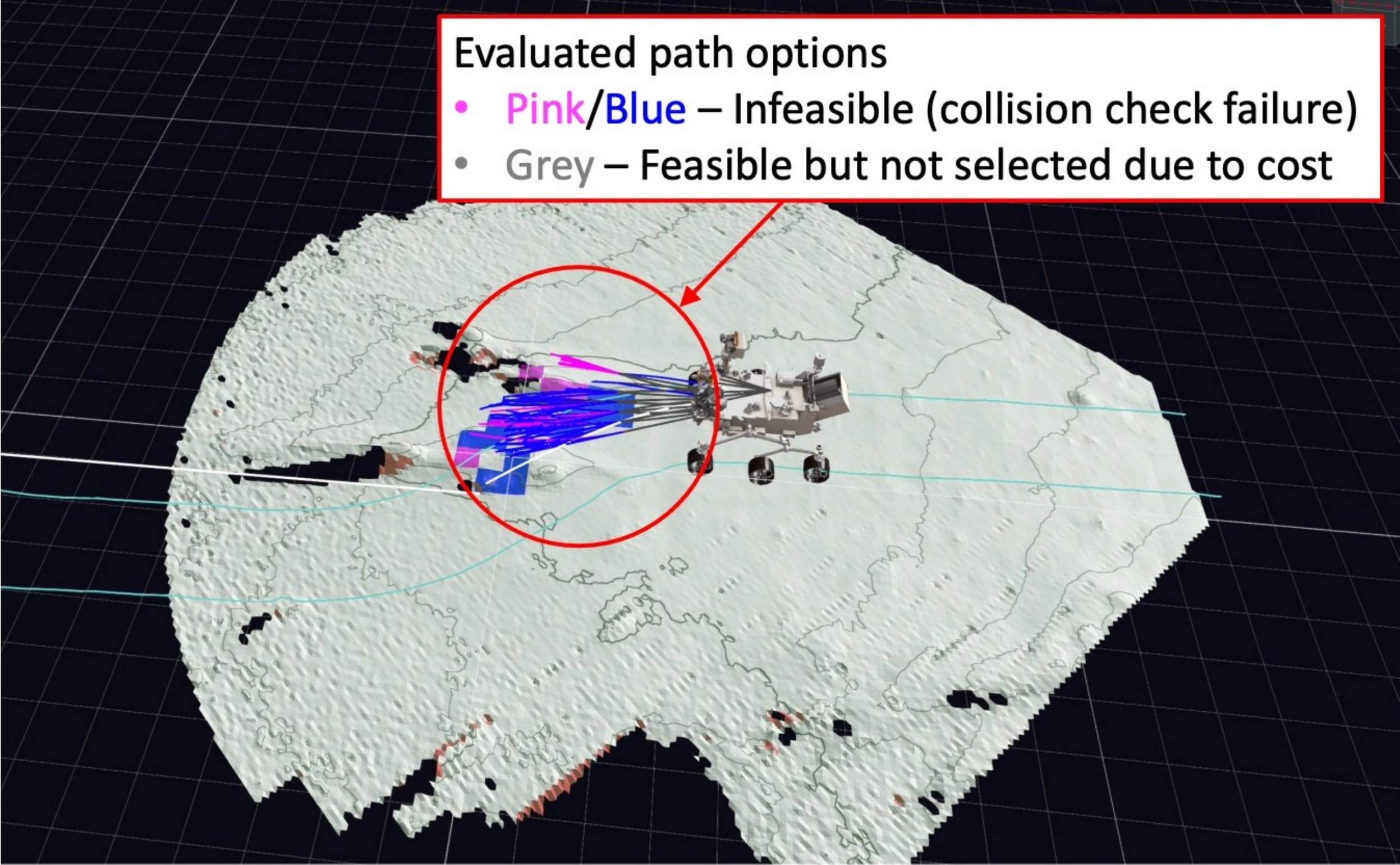}\\
    \vspace{-.27in}
    \includegraphics[width=\linewidth]{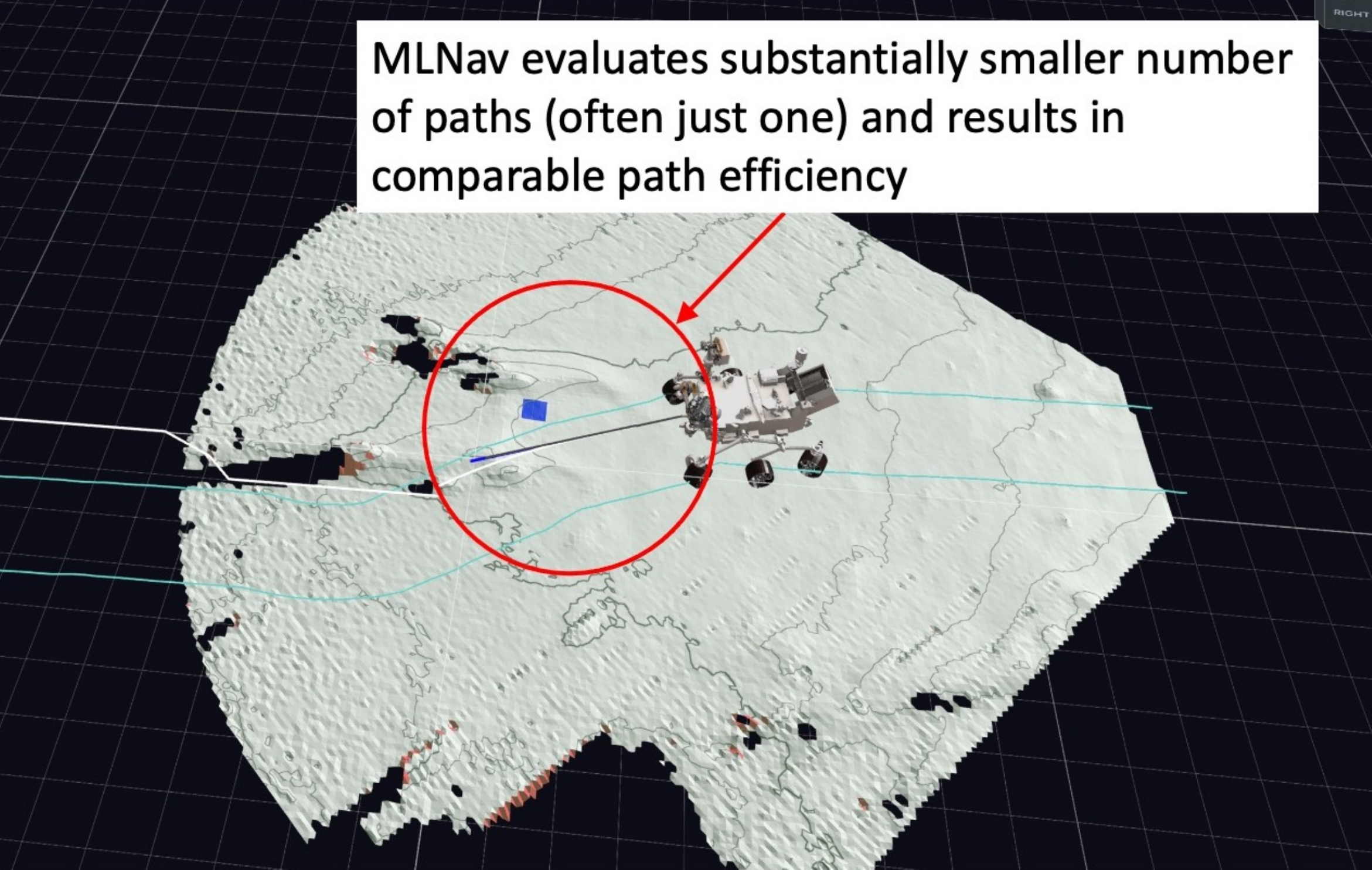}
    \vspace{-5mm}
    \caption{Paths planning on real Martian terrain (Perseverance Sol 122) with Baseline (ENav; top) and MLNav (bottom).}
    \vspace{-2mm}
    \label{fig:sol122-visualization}
\end{figure}

\subsection{Results}
The results on both Sols agree with the results on synthetic terrains presented in the previous section. 
Both ENav and MLNav found feasible paths to the goal as expected, and the paths are qualitatively similar. 
This is consistent with the result in Table \ref{tab:enav-ablation} that MLNav gives relatively minor improvement in path inefficiency, particularly on benign terrains.
Since both of Sol 122 and Sol 178 terrains fall under the "benign" category in Table \ref{tab:enav-ablation}, it is expected that the baseline algorithm can find a path as good as MLNav. 
Notable difference were observed when the rover avoided obstacles. 
For example, Figure \ref{fig:sol122-visualization} is the visualization of Sol 122 drives by ENav and MLNav when the rover was avoiding a rock. 
The blue and pink lines seen on the ENav visualizations are the paths on which ENav ran ACE evaluations and ended up with not choosing. 
Observe that there is only one blue line in the MLNav drive at the same location. 
This means that the second-ranked path based on ML heuristics turned out to be feasible, hence MLNav ran ACE only on two paths in this planning cycle. 
Of course, MLNav occasionally needed to ran ACE on many path options until finding a feasible one, but overall MLNav ran substantially smaller number of ACE collision checks and, in the majority of the planning cycles, it evaluates only a single path.
The complete movies of Sol 122 and 178 drives are attached as supplemental materials.

Table \ref{tab:mars-data-evaluation} shows the quantitative results. On both sols, there was only trivial changes in path inefficiency, indicating that MLNav resulted in a qualitatively similar path as discussed above.
In contrast, a substantial improvement in the number of ACE checks is observed. Again, this is because the top-ranked path evaluated by the ML-based heuristics was often feasible even in the presence of obstacles. 
While the number of test cases with real Martian terrain was limited for practical reasons mentioned above, this experiment demonstrated the ability of MLNav to improve the performance of path planning in a real environment.
This result is particularly remarkable because, as explained in Section \ref{sec:enav}, the training data that we used for this experiment was produced solely with synthetic terrains \textit{before} the landing of the rover. 

\begin{table}[!t]
    \caption{Path Planning Results on real Mars data}
    \label{tab:mars-data-evaluation}
    \centering
    \begin{tabular}{c c c c c}
    \toprule
    \\[-2mm]
     & \multicolumn{2}{c}{Sol 122}  & \multicolumn{2}{c}{Sol 178} \\
    \textbf{Metric} & \textbf{Baseline} & \textbf{MLNav} & \textbf{Baseline} & \textbf{MLNav}\\
    \\[-2mm]
    \midrule
    \midrule
    Path Inefficiency (\%) &3.1 & 2.0 & 0.13 & 0.66  \\
    \midrule
    Number of ACE Checks & 284 & 42.7 & 271 & 28.3  \\
    \midrule
    Overthink Rate (\%) & 8.3 & 4.2 & 5.6 & 0 \\
    \bottomrule
    \end{tabular}
\end{table}

%% file: sec_conclusion.tex
In this paper we presented \emph{MLNav} – a holistic framework for high-stakes planning that allows resource-constrained robotic systems to effectively navigate in complex environments while guaranteeing safety. Our main contribution was a general system design principle for effectively integrating ML methods into existing navigation pipelines of safety-critical robotic systems. We studied the efficacy of our framework through a concrete case study on Mars rover navigation, and demonstrated substantial improvements across several key performance metrics, using high-fidelity simulations with both real Martian terrain collected by the Perseverance rover and a suite of challenging synthetic terrains. In future work, we plan to further validate the performance of MLNav through end-to-end demonstration on an analog Mars rover.